%% file: main.tex
\documentclass[11pt,a4paper]{article}
\usepackage[hyperref]{resources/emnlp-ijcnlp-2019}
\usepackage{times}
\usepackage{latexsym}
\usepackage{graphicx}
\graphicspath{ {./} }
\usepackage{url}
\usepackage{xspace}
\usepackage{comment}

\aclfinalcopy 

\input{commands.tex}

\usepackage{listings}
\title{A Dataset of General-Purpose Rebuttal}

\author{
Matan Orbach$^1$,
Yonatan Bilu$^1$,
Ariel Gera$^1$,
Yoav Kantor$^1$, 
Lena Dankin$^1$, 
Tamar Lavee$^1$,
\\
\textbf{
Lili Kotlerman\thanks{\ \ This work was done at IBM Research.}$^{\ \ 2}$,
Shachar Mirkin$^{*3}$, 
Michal Jacovi$^1$,
Ranit Aharonov$^1$ 
and
Noam Slonim$^1$ 
} \\ 
$^1$IBM Research, $^2$Intuition Robotics, $^3$Digimind
\\
\{matano,yonatanb,arielge,yoavka,lenad,tamarl, michal.jacovi, ranita,noams\}@il.ibm.com,
\\
lili@intuitionrobotics.com, 
shachar.mirkin@digimind.com
}

\date{}

\begin{document}
\maketitle

\input{abstract}
\input{introduction}

\input{related_work}
\input{data.tex}
\input{experiments/experiments.tex}

\input{claim-matching.tex}
\input{Conclusions.tex}

\section{Acknowledgments}
We thank the anonymous reviewers for their valuable comments, and Hayah Eichler for creating the initial \knowledgeBaseName.

\bibliographystyle{resources/acl_natbib}
\bibliography{main}

\end{document}

%% file: commands.tex
\newcommand{\tableRef}[1]{Table \ref{#1}}
\newcommand{\sectionRef}[1]{\S \ref{#1}}
\newcommand{\figureRef}[1]{Figure  \ref{#1}}

\newcommand{\percentage}[1]{$#1$\%\xspace}
\newcommand{\statistic}[1]{$#1$\xspace}
\newcommand{\name}[1]{\emph{#1}\xspace}
\newcommand{\supp}[0]{Appendix\xspace}

\newcommand{\claim}[0]{claim\xspace}
\newcommand{\claims}[0]{claims\xspace}
\newcommand{\CRKBclaims}[0]{GP-claims\xspace}
\newcommand{\CRKBclaim}[0]{GP-claim\xspace}
\newcommand{\knowledgeBaseName}[0]{GPR-KB\xspace}
\newcommand{\datasetName}[0]{GPR-KB-55\xspace}

\newcommand{\action}[0]{[ACTION]\xspace}
\newcommand{\topic}[0]{[TOPIC]\xspace}

\newcommand{\wordTwoVecBaselineName}[0]{\name{w2v}}
\newcommand{\PriorName}[0]{\name{Prior}}
\newcommand{\priorName}[0]{\name{prior}}
\newcommand{\bertName}[0]{\name{bert}}
\newcommand{\idebateBaselineName}[0]{\name{w2v-i18-claims}}
\newcommand{\idebateDetailedBaselineName}[0]{\name{w2v-i18}}
\newcommand{\worTwoVecLeadsBaselineName}[0]{\name{w2v-GP-claims}}
\newcommand{\bertTestName}[0]{\name{bert-test}}
\newcommand{\bertBaselineName}[0]{\name{bert}}
\newcommand{\bertWithIdebateBaselineName}[0]{\name{bert+}}

\newcommand{\idebate}[0]{iDebate\xspace}

\newcommand{\idebateDatasetName}[0]{iDebate18\xspace}
\newcommand{\idebateSpeechCoverage}[0]{\percentage{86.5}}
\newcommand{\idebateAverageClaimsLabeledPerSpeech}[0]{\statistic{4.4}}
\newcommand{\idebateAverageMentionsInSpeeches}[0]{\statistic{1.8}}

\newcommand{\averageTokensInSpeech}[0]{\statistic{833.1}}
\newcommand{\averageSentencesInSpeech}[0]{\statistic{28.7}}

\newcommand{\numRebuttals}[0]{$55$\xspace}

\newcommand{\leadToMotionKappaTwoAnswers}[0]{$0.45$\xspace}
\newcommand{\leadToMotionKappaThreeAnswers}[0]{$0.52$\xspace}
\newcommand{\leadToMotionRelevant}[0]{\percentage{46}}
\newcommand{\leadToMotionSupport}[0]{\percentage{20}}
\newcommand{\leadToMotionContest}[0]{\percentage{26}}

\newcommand{\averageRelevantLeadsPerMotion}[0]{\statistic{25.4}}

\newcommand{\averageRelevantLeadsPerMotionLenient}[0]{\statistic{16.2}}

\newcommand{\numLeadToSpeechLabels}[0]{3,246\xspace}
\newcommand{\numLeadToSpeechCrowdLabelers}[0]{\statistic{22}}
\newcommand{\leadToSpeechMentioned}[0]{\percentage{41}}
\newcommand{\leadToSpeechExplicit}[0]{\percentage{7}}
\newcommand{\leadToSpeechImplicit}[0]{\percentage{34}}

\newcommand{\leadToSpeechAgreementBinary}[0]{\statistic{0.37}}
\newcommand{\averageMentionsPerSpeech}[0]{\statistic{6.7}}
\newcommand{\averageExplicitMentionsPerLead}[0]{\statistic{4.3}}
\newcommand{\averageImplicitMentionsPerLead}[0]{\statistic{17.6}}



%% file: abstract.tex
\begin{abstract}
In Natural Language Understanding, the task of response generation is usually focused on responses to short texts, such as tweets or a turn in a dialog. Here we present a novel task of producing a critical response to a long argumentative text, and suggest a method based on general rebuttal arguments to address it.
We do this in the context of the recently-suggested task of listening comprehension over argumentative content: given a speech on some specified topic, and a list of relevant arguments, the goal is to determine which of the arguments appear in the speech. 
The general rebuttals we describe here 
(written in English) 
overcome the need for topic-specific arguments to be provided, by proving to be applicable for a large set of topics. This allows creating responses beyond the scope of topics for which specific arguments are available.
All data collected during this work is freely available for research\footnote{\url{https://www.research.ibm.com/haifa/dept/vst/debating_data.shtml}}.
\end{abstract}

%% file: introduction.tex
\section{Introduction}

A key element in argumentation is rebuttal: 
the ability to contest an argument by presenting a counter-argument.
It is an important skill, not easily learned, and valued in many fields such as politics and science. It is useful for advancing your own views and beliefs over opposing ones, but perhaps more importantly, it facilitates a critical examination of the views and beliefs that you hold.
An automatic rebuttal system could therefore be useful whenever critical analysis of written or spoken content is required -- be it an elementary school student writing an essay or a seasoned journalist composing an op-ed.

In the context of Natural Language Understanding, the study of rebuttal and counter-arguments has focused on elucidating such relations between given arguments. Indeed, such ``attack'' relations are the foundation of {\em Argumentation Frameworks} \cite{dung95}; such frameworks have been one of the main objects of study in computational argumentation.

A related task, that of generating a response which need not be a rebuttal or even argumentative, has been the subject of much research, especially in the context of dialog systems, chat bots, and question answering. 
In this line of work the response typically follows a short input text, often only a sentence or two.

Here we suggest the task of producing a rebuttal in response to a long argumentative text. Specifically, we consider spoken speeches around four minutes long. In addition to being longer, and perhaps because they are so, these kinds of texts tend to include very general claims, often implicit in the text. As such, these claims may appear in varied contexts, and it may be feasible to compile a list of such claims independently of the speeches' topics.

For example, a concern that often comes up in debates about policy is that implementing the policy (or failing to do so) disproportionately harms minorities. This claim can be made to oppose school vouchers, to oppose voter registration laws or to support criminal justice reform.
Moreover, for some such general claims, it is feasible to phrase a single rebuttal response which can fit many of the contexts in which the claim might be made. In the above example, such a response may talk about separating between the policy at hand, which should be adopted based on its merits, and the need to right historical wrongs, which should be pursued independently.

We envision an automatic rebuttal system based on this observation, which includes a manually curated General-Purpose Rebuttal Knowledge Base (\knowledgeBaseName) comprised of claims and matching rebuttal responses.
Given an argumentative text, the system would identify which \claims from the \knowledgeBaseName are made in the text (explicitly or implicitly), and 
produce a rebuttal using the available counter-arguments.
Clearly, many of the claims made in the text would not appear in the \knowledgeBaseName. 
The objective is therefore not to identify and rebut all arguments, but rather to identify and rebut {\em some} arguments, and construct a \knowledgeBaseName that facilitates that.

Such a system (based on the more elaborate CoPA modeling of \citealp{bilu2019argument}) was indeed implemented as a key element in IBM's Project Debater rebuttal mechanism, and demonstrated during the live debate held between it and debating champion Harish Natarajan\footnote{Video of the debate at \url{https://www.youtube.com/watch?v=m3u-1yttrVw}}. However, a rebuttal system of this nature may be of interest beyond the realm of debating technologies. For example, such a system may be instrumental in making media consumption a more critical process, by automatically challenging the consumer with counter-arguments. Similarly, it can be applicable in the education domain, stimulating critical thinking by prompting students with counter-arguments in response to (or during) essay-composition tasks.

The formation of a \knowledgeBaseName that is applicable to the real world poses several challenges.
First, phrased \claims must be both relevant to a variety of topics, and
commonly used.
Second, pre-written rebuttals should be effective and persuasive, even though they are created without prior knowledge of the context in which they are to be of use. 
We address these issues by turning to a domain in which a similar problem is solved by humans: the world of competitive debates.
In these contests, successful debaters need to combine specific knowledge about the topic at hand, with general arguments that arise from the underlying principles of the debate. 
Their ability to use such general arguments for different topics lays the basis for using a \knowledgeBaseName as the one described above.

Accordingly, we asked an expert debater to create the initial \knowledgeBaseName by suggesting common \claims and preparing matching rebuttals.
The full process is detailed in \sectionRef{subsec:CRKB_construction}.

To assess the usefulness of the suggested claims and rebuttals in the real world, we performed several steps of labeling on the dataset we constructed in \cite{Mirkin-etal:idebate}, containing spoken argumentative content discussing controversial topics. 
Details of this process, along with an analysis showing the high coverage obtained by our knowledge base, are described in \sectionRef{sec:experiments}.

Another major challenge is the development of automatic methods for identifying whether knowledge-base claims are mentioned by speakers.
We break this problem into a three-stage funnel -- identifying whether: (i) a claim is relevant to the topic; (ii) the claim's stance aligns with that of the speaker; (iii) the claim was made by the speaker. We provide simple baseline results for this third step (\sectionRef{sec:detection}). Interestingly, we observe that simply selecting the claim with the highest acceptance rate in the training data (without looking at the text) provides a challenging baseline.

The main contributions of this work are (i) the introduction of a novel task in NLU: producing rebuttal in response to a long argumentative text (ii) a manually constructed \knowledgeBaseName shared across multiple topics (iii) an additional layer of labeling to our dataset from \citet{Mirkin-etal:idebate} for such claims (iv) a baseline for detecting whether such a claim was mentioned in a speech.

%% file: related_work.tex
\section{Related Work}
In \citet{Mirkin-etal:idebate} we introduced the task of Listening Comprehension over argumentative content. 
That work analyzes recorded speeches, and tries to identify whether arguments from iDebate\footnote{https://idebate.org/debatabase} are mentioned in the speech. 
Similarly, in \citet{Lavee2019towardsRebuttal} we addressed this task by first mining arguments from a large news corpus, and then identifying the arguments which are mentioned in speeches.

This work complements our previous works in two ways. First, the \knowledgeBaseName constructed here is of general claims, with wide cross-topic relevance.
It facilitates Listening Comprehension for topics not mentioned in iDebate, or topics for which automatic argument mining does not yield satisfactory results. 
Second, while \citet{Mirkin-etal:idebate} mention that the iDebate counter points can in principle be used for rebuttal, and \citet{Lavee2019towardsRebuttal} suggest mining opposing arguments from their corpus to counter arguments mentioned in speeches, pursuing both ideas is left for future work.
We pick up the baton (in the context of the \knowledgeBaseName suggested here), and annotate the validity of the counter arguments as rebuttal to the ideas expressed in a matching speech.

Response generation has been the subject of much research, using a wide variety of methods (e.g. \citealp{ritter2011data, sugiyama2013open, shin2015context, yan2016learning, xing2017topic}). 
In the context of dialog systems (see recent survey in \citealp{chen2017survey}), there is usually a distinction between task-oriented systems \cite{DBLP:journals/corr/WenTaskOrientedDialog} and open-domain ones \cite{DBLP:journals/corr/Mazare2018PersonalizeDialogAgents, Weizenbaum1966eliza}. The task here can be seen as lying in between the two: on the one hand it allows for a response to speeches on a variety of topics; on the other, the response is restricted to be a rebuttal of a claim made in the speech. A major difference from dialog systems is that in this task the analysis is of a complete speech - rather than taking turns, and the goal is to respond to some of the claims - but not necessarily all.

In the context of computational argumentation much attention has been given to mapping rebuttal or disagreement among arguments. Such works include datasets exemplifying these relations \cite{walker2012corpus, peldszus2015annotated,  musi2017building}, modeling them \cite{sridhar2015joint} and explicitly detecting them \cite{rosenthal2015couldn, peldszus2015towards, wachsmuth-etal-2018-retrieval}. The \knowledgeBaseName in this work is reminiscent of argument datasets that depict rebuttal relations, but the arguments are of a different type, being manually authored as general and applicable to a wide range of topics.

Most similar to our work is the task of generating an argument from an opposing stance for a given statement \cite{hua-acl2018-rebuttal, hua-acl2019-rebuttal-generation}. These works present a neural-based generative approach, 
and experiment with user-written posts.
Our task differs in that the input is longer text, potentially containing multiple arguments.

%% file: data.tex
\section{Data}

\input{tables/claim_rebuttal_examples.tex}

\subsection{Motions and Speeches}
The speeches analyzed in this work are the 200 speeches provided by \citet{Mirkin-etal:idebate}. 
Each speech debates one of 50 motions originating from \idebate. 
In this data, the phrasing of the motions is often simplified 
to include an explicit {\em topic} and {\em action}. 
For example, the \idebate motion {\em This House would introduce goal line technology in football} is simplified to {\em We should introduce goal line technology}, where the topic is {\em goal line technology} and the action is {\em introduce}.

Speeches are evenly distributed between motions, each having two speeches supporting it (i.e. the speaker is arguing in favor of the motion) and two contesting it. 
They were recorded by $14$ different speakers.
A speech is given in several formats. 
We use the recorded audios and manually-created transcriptions.
Recordings are about $4$ minutes long,
and the transcript texts contain on average \averageSentencesInSpeech sentences and 
\averageTokensInSpeech tokens.

Lastly, the dataset contains claims taken from \idebate along with annotations identifying specific claims mentioned in particular speeches.
Herein we refer to this data as \idebateDatasetName.

\subsection{Knowledge base construction}\label{subsec:CRKB_construction} 
An experienced competitive debater was solicited to author claims that tend to come up in debates across varied topics, and to write a rebuttal argument for each such claim (see the \supp for the guidelines).
She was \emph{not} given access to any of
the \idebateDatasetName motions, which are analyzed later on.
In total, 39 pairs were constructed in this way.

Texts were allowed to incorporate the special tokens {\em \action} and {\em \topic}, which are replaced by the debate topic and suggested action when applied to a specific motion or speech. For example, in the context of the motion {\em We should introduce goal line technology}, the claim {\em \action\topic will encourage better choices} is translated to 
{\em introducing goal line technology will encourage better choices}.

In a second phase, the claim-rebuttal pairs were edited by the authors, as follows:

(i) Some rebuttal texts were written with the context of a full speech in mind, and included segments that refer to what a debater would include in such a speech. For example, one 
included the segment "I have proven that this method is effective". Such segments were edited out.

(ii) For some claims, it seemed that an opposite claim could also be made. In these cases the negation of the claim was also added to the knowledge base, along with an appropriate rebuttal. For example, in addition to the claim {\em "\action\topic is the most practical way to solve the problem."}, we also added the claim {\em "\action\topic is not the most practical way to solve the problem."}.

After these modifications, the final knowledge base includes \numRebuttals claim-rebuttal pairs.
Claims are always one short single sentence, with an average length of $8.5$ tokens.
Rebuttals are longer, on average $1.8$ sentences long, and containing on average $32.9$ tokens.
Three examples from the \knowledgeBaseName are given in \tableRef{tab:claim_rebuttal_examples}; henceforth we refer to the generated claims as \emph{\CRKBclaims}, or simply as {\em claims} when the context is clear.

%% file: tables/claim_rebuttal_examples.tex
\begin{table*}[t]
\begin{center}
\begin{tabular}{|p{0.24\linewidth}|p{0.71\linewidth}|}

\hline \bf Claim & \bf Rebuttal \\ \hline
We must limit personal choice in this case &
The greater good means nothing if the rights of individuals are being violated. It doesn't make sense to violate rights in order to protect them.
\\
\hline
\action\topic is good for the economy & 
While we need to take the economy into account when making decisions, it cannot be the sole consideration or even the top priority in many cases. In this case, the harms outweigh any benefits there may be to the economy.
\\
\hline
We need to protect the weakest members of society &
A truly fair society is one where different people are afforded similar rights and are also trusted to look after themselves. While weaker segments of society can be more vulnerable, this does not justify paternalistic policies that are not beneficial for society as a whole.
\\
\hline
\end{tabular}
\end{center}
\caption{Examples of \CRKBclaims and matching rebuttals, created through the process described in \sectionRef{subsec:CRKB_construction}.
}
\label{tab:claim_rebuttal_examples} 
\end{table*}

%% file: experiments/experiments.tex
\section{Annotation Experiments}\label{sec:experiments}

Four annotation experiments are described next, aimed at assessing the applicability of the generated \knowledgeBaseName to the real world.
Each of the following subsections describes one experiment and its results.
An overview of the whole process is depicted in Figure \ref{fig:annotation-summary}.
The full annotation guidelines for each experiment appear in the \supp.
\input{figures/annotation_summary.tex}

\input{experiments/relevancy-to-motion.tex}
\input{experiments/usage-spoken-content.tex}

\input{experiments/sentence-annotation.tex}
\input{experiments/rebuttal-verification.tex}

\subsection{The \datasetName Dataset}
The annotation results show that it is possible to construct a concise set of general claims, such that in most speeches at least one 
of them 
will come up. Furthermore, they show that a rebuttal to these claims can be authored independently of the specific motions and speeches, while nonetheless being a plausible response in their context.
Table \ref{tab:annotation-stats} summarizes the statistics for the pair-annotation experiments.
\input{tables/annotation_stats.tex}
The resulting dataset is freely available, and is one of the main contributions of this work\footnote{\url{https://www.research.ibm.com/haifa/dept/vst/debating_data.shtml}}. 
We name the new dataset \datasetName.

%% file: figures/annotation_summary.tex
\begin{figure}[t]
\centering
\includegraphics[trim={1cm 1cm 1.5cm 0.5cm},clip,width=75mm]{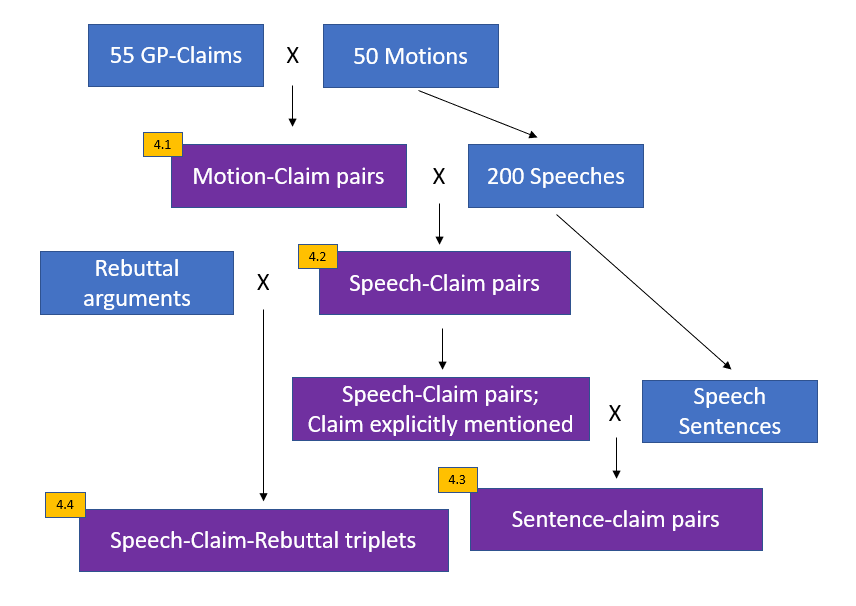}
\caption{
Annotation overview: 
All motion-claim pairs were annotated for whether the claim is relevant to the motion (see \sectionRef{subsec:leadToMotions}).
For each claim,
speeches discussing the relevant motions 
were annotated for whether the claim was mentioned in the speech (see \sectionRef{subsec:mention}), explicitly or implicitly.
For explicitly mentioned claims,
selected speech sentences 
were annotated for whether the claim was mentioned in the sentence (see \sectionRef{subsec:sentencemention}). 
In addition, for claims mentioned in the speech, the corresponding Rebuttal Argument was annotated for whether it is a plausible rebuttal in the context of the speech (see \sectionRef{subsec:rebuttalValidation}). 
Blue rectangles indicate textual resources, violet ones indicate annotated resources, yellow ones refer to the relevant subsection.\label{fig:annotation-summary}
}
\end{figure}

%% file: experiments/relevancy-to-motion.tex
\subsection{Cross-Topic Relevancy}\label{subsec:leadToMotions}

The \CRKBclaims were written based on the experience of a professional debater, but without context of specific topics. 
The first annotation experiment aims to establish whether these claims indeed attain the desired goal of being applicable to a varied set of topics.
For each motion in \idebateDatasetName, and for each \CRKBclaim, we asked annotators
to decide whether the \claim supports the motion, opposes it or is not relevant\footnote{The stance is required for the experiment in \sectionRef{subsec:mention}.}. 
Annotation was done by $7$ experienced annotators, and $5$ answers were collected for each question.

A \CRKBclaim was considered \emph{relevant} to a motion when marked as supporting or opposing it by most annotators.
The stance of relevant \claims towards the motion was determined by majority. 
When a relevant claim has an equal number of supporting and opposing answers, its stance is considered undetermined.

\paragraph{Results} Annotation included 2,750 claim-motion pairs\footnote{All pairs of $50$ motions and $55$ claims.}, of which \leadToMotionRelevant are \claims annotated as relevant to the motion. \leadToMotionSupport are annotated as supporting the motion, \leadToMotionContest as opposing it, and a negligible number have an undetermined stance.

On average, \averageRelevantLeadsPerMotion claims are annotated as relevant per motion.
Inter-annotator agreement (average Cohen's Kappa \citep{cohenKappa} over pairs of annotators), is \leadToMotionKappaThreeAnswers for the three-labels task, and \leadToMotionKappaTwoAnswers for the binary label of relevant/irrelevant.

\input{figures/leadToMotionHistogram.tex}

\figureRef{fig:lead-to-motion-histogram} shows the distribution of \claims vs. the number of motions annotated as relevant.
Of note, many \claims are relevant to various motions: \percentage{56} (sum of the four right-most bars)
are relevant to at least $20$ (out of $50$) motions. However, only \percentage{5} are phrased in a manner so general that they may be relevant to all $50$ motions. 
An example of such a general claim is \emph{\action \topic will harm others}.
At the other extreme, the claim 
\emph{Animals have rights} is labeled as relevant to only $3$ motions discussing various animal-related issues: adopting vegetarianism, banning bullfighting and legalizing ivory trade.

The majority of the claims are therefore general enough to be relevant to a substantial portion of the motions, but not so general as to make them trivially relevant to all motions.

%% file: figures/leadToMotionHistogram.tex
\begin{figure}[t]
\centering
\includegraphics[trim={2cm 9cm 1.5cm 9cm},clip,width=75mm]{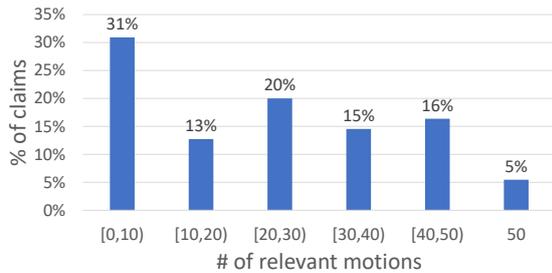}
\caption{
The distribution of \CRKBclaims vs. the number of motions annotated as relevant.\label{fig:lead-to-motion-histogram}
}
\end{figure}

%% file: experiments/usage-spoken-content.tex
\subsection{Usage in Spoken Content}\label{subsec:mention}

Having established that \CRKBclaims are potentially relevant to many motions, the question still remains of whether or not they are actually commonly made by people debating these motions. This is a crucial point for using them as a basis for generating a rebuttal-response. 

To assess this, 
annotators were shown speeches 
from \idebateDatasetName, alongside a matching list of \CRKBclaims determined to be relevant in the previous stage. 
Specifically, claims annotated as supporting a motion were shown for speeches in which the speaker is arguing in favor of that motion, and vice versa. To allow for a greater number of potential claims, those which at least $2$ annotators considered relevant (rather than $3$) were included. Claims with an undetermined stance were excluded.

Speeches were presented in both audio and text formats, 
and annotators were allowed to choose between listening, reading or both.
They then had to determine whether each claim is mentioned in the speech {\em explicitly}, {\em implicitly} or not at all.
The number of claims presented for each speech was limited to $20$; in case a larger number was determined to be relevant, the question was split 
into chunks of $20$ claims.

In total, 
\numLeadToSpeechLabels claim-speech pairs required annotation, almost four times more than the corresponding annotation included in \idebateDatasetName. 
Annotation was done 
using the \textit{Figure-Eight}\footnote{\url{www.figure-eight.com} (formerly CrowdFlower).} crowd-sourcing platform, with $10$ annotators per question. 
Clearly, this is a challenging task for the crowd, and hence a selected group of \numLeadToSpeechCrowdLabelers annotators was used. Selection was based on their past performance on other tasks done by our team. 

To further validate the annotation, the list of claims presented for each speech included claims for which the correct label was known a-priori.
These include claims annotated in the previous experiment as irrelevant for the motion, for which it is assumed that the correct label is ``not mentioned''.
In addition, annotation was done in batches.
Claim-speech pairs for which unanimous answers were obtained in earlier batches were included in newer ones, with the correct label assumed to be this unanimous answer.

A claim is considered \emph{mentioned} in a speech if a majority labeled it as mentioned (i.e. summing up implicit and explicit answer counts). Otherwise it is considered as \emph{not mentioned}.
A mentioned claim is \emph{explicit} in the speech if its explicit answers count strictly exceeds its implicit answers count.
Otherwise, it is considered \emph{implicit}. 

\paragraph{Results} \leadToSpeechMentioned of claim-speech pairs were labeled as mentioned (\leadToSpeechExplicit explicit and 
\leadToSpeechImplicit 
implicit).
On average, each \claim is explicit in \averageExplicitMentionsPerLead speeches, and implicit in \averageImplicitMentionsPerLead speeches. 

Pairwise inter-annotator agreement is \leadToSpeechAgreementBinary when considering 
two labels: 
mentioned or not.
The average error rate of all annotators, on questions with a prior known answer, is \percentage{7}, suggesting a relatively high-quality annotation.

The \emph{prior} of a claim is defined as the number
of speeches in which it was found to be mentioned, divided by the number of speeches in which it was labeled.
\figureRef{fig:lead-to-speeches-histogram} depicts the percentage of claims vs. their prior, separately for explicit and implicit mentions.
Some claims are never mentioned in any speech: \percentage{20} are never mentioned explicitly and $\sim$\percentage{10} are not mentioned at all, not even implicitly. 
Note that for the most part, claims are mentioned in less than half of the speeches for which they may be relevant. 

In conclusion, 
the 
results suggest that \CRKBclaims are often used in spoken content discussing various topics, and that this is not due to a small subset of trivial claims. Rather, most claims appear at least once, but usually in no more than half of the speeches for relevant motions.

These properties make automatic detection of these claims in speeches an interesting and challenging task.

\paragraph{Comparison to \idebateDatasetName}
\tableRef{tab:leadToSpeech:comparisonToIdebate} compares the results of this annotation to that of \idebateDatasetName, which
contains topic-specific claims annotated for the same set of speeches\footnote{Available answers in \idebateDatasetName are \emph{mentioned} or \emph{not mentioned.}}.

Surprisingly, topic-specific claims are no more likely to occur in speeches discussing that particular topic.
Moreover, the larger number of potential \CRKBclaims leads to a higher absolute number of mentions, to the extent that -- in contrast with \idebateDatasetName { --} all speeches include at least one mention.
Hence, the \knowledgeBaseName augments the \idebateDatasetName dataset, both by 
increasing the number of claims that are to be sought in a speech, and by suggesting claims to speeches for which \idebateDatasetName does not contain any. 

\input{tables/comparisonToIdebate.tex}

\input{figures/leadToSpeechHistogram.tex}

%% file: tables/comparisonToIdebate.tex
\begin{table}[ht!]
\begin{center}
\begin{tabular}{|l|c|c|}
\hline \textbf{Speech stat.} & \textbf{\CRKBclaims} & \textbf{\idebateDatasetName}
\\ \hline
 Coverage & \percentage{100}  & \idebateSpeechCoverage
 \\ \hline
 Avg. Mentions & \averageMentionsPerSpeech  & \idebateAverageMentionsInSpeeches
 \\ \hline
 Avg. Potential & \averageRelevantLeadsPerMotionLenient  & \idebateAverageClaimsLabeledPerSpeech
 \\ \hline
\end{tabular}
\end{center}
\caption{A comparison of \CRKBclaims and topic-specific \idebate claims annotation.
Coverage is the percentage of speeches with at least one claim annotated as mentioned.
Avg. Potential is the average number of possibly relevant claims per speech, and Avg. Mentions is the average number of claims annotated as mentioned.}
\label{tab:leadToSpeech:comparisonToIdebate} 
\end{table}

%% file: figures/leadToSpeechHistogram.tex
\begin{figure}[t]
\centering
\includegraphics[trim={2.2cm 4cm 2.2cm 3.8cm},clip,width=75mm]{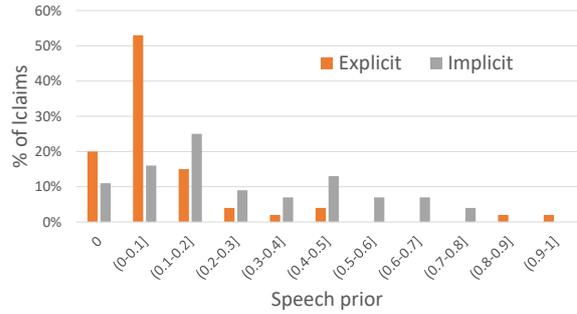}
\caption{
The distribution of \CRKBclaims vs. speech prior (the percentage of labeled speeches in which a claim is mentioned), for explicit or implicit mentions.
\label{fig:lead-to-speeches-histogram}
}
\end{figure}

%% file: experiments/sentence-annotation.tex
\subsection{Where was it said?}\label{subsec:sentencemention}
A straightforward approach to determining whether a claim was mentioned in a speech is to go over 
its
sentences, one by one, and decide whether the claim is equivalent to a sentence or implied by it, as indeed is done in \citet{Mirkin-etal:idebate}. This is a challenging task since, as described above, most mentions are implicit. In many cases one can not point to 
a single sentence 
mentioning the claim,
as the claim is implied by the general stance of the speaker. 

Even when a sentence does imply a claim, automatically inferring that 
may be hard. For example, for the motion \emph{We should end cheerleading}, a 
relevant opposing claim 
is \emph{Ending cheerleading limits personal choice}. We 
identified a sentence implying it, 
\emph{The only clear moral system we can derive is one in which we value individual preference}, yet it seems hard to deduce this automatically without considering the surrounding context which clarifies that the argument is about personal choice.

An annotation task for 
identifying \emph{where} a claim was mentioned is considerably more difficult than the aforementioned annotations. 
Determining ground truth is far from trivial, as annotators may point to different sentences within the same argument as being the location of the claim.

Nonetheless, such information seems a valuable part of a GPR-KB. To
provide at least a partial solution, we annotated claim-sentence pairs directly,
asking whether the claim is mentioned within the sentence. 
Algorithms developed on such data
can then predict a claim as mentioned in a speech when it is mentioned in one of its sentences.
This form of annotation is simple and facilitates easier collection of a large number of labels.
To enable research in this direction, such annotations were performed both for \CRKBclaims and, since none are provided in \idebateDatasetName, for claims from \idebate.

A careful selection of pairs to annotate is required since there are too many pairs for a comprehensive labeling, and sampling at random would rarely yield a pair such that the claim is mentioned in the sentence.
Thus, we limited annotation to claims which were labeled as mentioned explicitly (and assumed all \idebate claims to be so), and paired them only with sentences which are somewhat similar to them (based on word2vec, \citealp{mikolov2013w2v}). 
Annotation was again done by $10$ crowd annotators.

\paragraph{Results}
Annotation included 4,271 \CRKBclaim and sentence pairs and 2,164 \idebate-claim and sentence pairs, with a similarity of at least $0.5$ and $0.7$ (resp.) between claim and sentence.
The usage of general crowd required 
some quality control.
Annotators not meeting one or more of the following criteria were removed, along with their answers: Answer at least $10$ questions; have at least $5$ common answers with $3$ different peers; have average agreement with peers $\geq{0.2}$.
 
The resulting inter-annotator agreement was $0.55$ for \CRKBclaims and $0.46$ for \idebate claims. 
Considering only pairs with at least 5 remaining answers, after filtering out annotators as described above, \percentage{20} of the \CRKBclaims-sentence pairs were annotated as a match (and \percentage{17} of \idebate pairs).

%% file: experiments/rebuttal-verification.tex
\subsection{Validity of Rebuttal Arguments}\label{subsec:rebuttalValidation}
Recall that rebuttal arguments in the \knowledgeBaseName were written without any specific contexts in which they are to be used.
Hence, 
even if the claim they respond to is indeed mentioned in a speech, it is not clear whether 
the pre-written rebuttal 
would consist a plausible rebuttal response to the speech.

We assessed the effectiveness of the rebuttal arguments 
using a two-step procedure.
First, as in \sectionRef{subsec:mention}, annotators were shown a speech and a claim, 
and determined whether the claim is mentioned in the speech.
Then, if they marked that claim as mentioned, its pre-written rebuttal 
was shown, and they were asked whether it is a plausible response to the mentioned claim \emph{in the context of the speech}.

This two-step annotation procedure was chosen for three reasons. 
First, requiring annotators to assess whether a claim is mentioned in a speech motivates them to review its content again and locate the relevant parts in which the claim is expressed. 
Second, it prevents irrelevant answers from those who do not think that the claim is mentioned in the speech.
Finally, asking again about a claim's mention enables result validation, when the answer is known a-priori with high confidence. 
Specifically, claims for which the annotation is unanimous were used for this purpose.

For each claim, we annotated two randomly sampled speeches mentioning it.
This amounted to $103$ rebuttal-speech pairs, since not all $55$ \CRKBclaims were mentioned in two speeches. 
We relied on the same group of crowd annotators who took part in the previous experiment, and once again required $10$ answers for each question. 

\paragraph{Results} 
Measuring agreement using Cohen's Kappa in this scenario is problematic. First, the label distribution is very biased: If rebuttal arguments are always plausible responses, then the correct answer is always yes. Any deviation from that will greatly reduce the score, making it an ill-fitting measure for such data \cite{jeni2013facing}. Second, the 
small number of questions leads to many annotator-pairs whose set of common questions, on which this score is computed, is rather small. This makes the computation unstable, 
since when averaging over all pairs, such small intersections contribute as much as the large ones.

Instead, looking at the majority annotation for each rebuttal argument, we observed that in \percentage{87} of the cases it indicates that the rebuttal is plausible. This suggests that regardless of whether annotators agree with one another, they tend to agree that the rebuttal is usually plausible. 
Moreover, computing the average kappa 
between annotators and 
the majority annotation, and considering only those annotators who answered at least $20$ questions, yields 
$0.47$. This value for such biased data, alongside an average error rate of \percentage{4} on the 
questions with known answers, suggests that the annotation is of reasonable quality.

\paragraph{Analysis} The results above show most rebuttals are appropriate in the vast majority of contexts. 
We therefore decided that continuing with this costly annotation is not needed.
Furthermore, manual analysis of cases in which the rebuttal was unanimously found inappropriate showed this stemmed from the rebuttal being inappropriate for the topic, rather than 
a specific speech. 
For example, when discussing goal-line technology, in response to the \claim \emph{"introducing goal-line technology will lead to greater problems in the future"}, the pre-written rebuttal is \emph{"Governments have an obligation to their citizens in the here and now. The better off society is today, the more resources we will have to make the future better when it comes"}. 
Such a response makes several assumptions which 
are clearly violated here,
such as the involvement of government.
Thus, further validation of rebuttals may benefit from first verifying their relevancy to the topic.

%% file: tables/annotation_stats.tex
\begin{table}[ht!]
\begin{center}
\begin{tabular}{|l|c|c|}
\hline \textbf{Annotated pairs} & \textbf{\# annotated} & \textbf{\# positive}
\\ \hline
 Motion -- \CRKBclaim & 2,750  & 1,265
 \\ \hline
 Speech -- \CRKBclaim & 3,246  & 1,491
 \\ \hline
 Sentence -- \CRKBclaim & 4,271  & 854
 \\ \hline
 Sentence -- & &  \\
 \idebate claim & 2,164  & 368
 \\ \hline
\end{tabular}
\end{center}
\caption{Summary of annotation experiment results. Positive examples are those in which a majority of the annotators indicated that the claim is relevant (motion) or mentioned (speech, sentence).}
\label{tab:annotation-stats} 
\end{table}

%% file: claim-matching.tex
\section{Detecting claims in speeches}\label{sec:detection}

Next, we establish baseline results for determining whether a \CRKBclaim is mentioned in a speech, and compare them to results 
obtained for \idebate claims.
For a fair comparison of the two data sources, we assume for both prior knowledge as to which claims are relevant for a motion, as well as their stance towards it. 
Hence, we take the labeling of \CRKBclaims to motions (described in \sectionRef{subsec:leadToMotions}) as given.
The following algorithms are considered:

\paragraph{word2vec} 
The best performing baseline of \citet{Mirkin-etal:idebate} utilizes a detailed description of each \idebate claim, comprised of several sentences.
It examines the speech sentence by sentence, and for each sentence computes its tf-idf weighted word2vec (\wordTwoVecBaselineName) similarity to the detailed claim description. 
A claim is then scored by taking the maximum over all claim-sentence similarity scores.
We use this method (\idebateDetailedBaselineName) as a baseline for the \CRKBclaims as well, yet sentences are scored by their similarity to the \CRKBclaim text, since no detailed topic-specific description exists. The latter is referred to as \worTwoVecLeadsBaselineName.
For comparison, we repeat the experiment using only the \idebate claim texts (\idebateBaselineName).

\paragraph{Bert} Recently, the Bert architecture \cite{devlin2018bert} has proven 
successful on similar tasks, and we provide its results as an additional baseline. Specifically, we select at random \percentage{80} of the motions as an ad-hoc train set, and the remainder as a test set (\bertTestName). 
Bert was trained on labeled claim-sentence pairs corresponding to motions from the train set, in two settings, considering: (i)  \CRKBclaims ($\sim$3K pairs) -- \bertBaselineName, and (ii) both \CRKBclaims and \idebate claims (almost 5K pairs) -- \bertWithIdebateBaselineName.
In inference, given a claim and a speech, sentences semantically similar to the claim (as in \sectionRef{subsec:sentencemention}) are 
scored by the fine-tuned network.
Their maximum is the outputted claim-speech score.

\paragraph{Prior} One important difference between \CRKBclaims and \idebate claims is that the same \CRKBclaim can (and does) appear across different motions and speeches. 
Specifically, given a training set, the a-priori probability that a \CRKBclaim will be mentioned in a speech can be computed.
Then, test claims are scored with their computed a-priori probability {\em without considering the text of the speech}.
This baseline is referred to as \priorName.

\paragraph{Results}
\figureRef{fig:precision-recall-comparison-to-idebate} plots precision-recall curves comparing claim detection baselines over \idebate claims and \CRKBclaims.
As observed by \citet{Mirkin-etal:idebate}, \wordTwoVecBaselineName
works 
best when given a detailed \idebate claim description.
Without it,
performance is comparable for the two claim sources, and is rather poor for both.
\PriorName results were obtained by using a leave-one-motion-out cross validation: at each fold a single motion is left out and the others are used for training.
Its precision-recall curve shows that when considering these statistics, it presents a challenging baseline.

\input{figures/precision_recall_compare_to_idebate.tex}

To 
compare the \bertBaselineName baseline to
others, the precision-recall curves for both \priorName\footnote{Here, a-priori probabilities were computed on all motions not in \bertTestName, then applied to motions of \bertTestName.} and \wordTwoVecBaselineName were computed over 
speeches from \bertTestName.
As shown in \figureRef{fig:precision_recall_ont_test}, while \bertName clearly outperforms \wordTwoVecBaselineName, it nonetheless does not reach the \priorName baseline.
The additional data provided in training to \bertWithIdebateBaselineName does not help.

Note that this comparison is between methods which are derived from different types of data. Here \bertName is trained only on explicitly-mentioned claims, with respect to (ostensibly) semantically similar \textit{sentences}. On the other hand, the \priorName baseline is computed based on all claims, and their annotation w.r.t. \textit{the entire speech}. This may be part of the reason why \bertName, which has proven to be successful on many NLP tasks, here achieves lower performance than this simple baseline.

\paragraph{Analysis} 
Although \priorName seems like a strong baseline 
in terms of precision and recall, it is probably not a desired solution by itself, since it simply produces high probability responses regardless of the rebutted content. 
For example, among its top-\percentage{20} predictions, precision is \percentage{83} and recall is \percentage{40}, yet they include only \percentage{22} of available \CRKBclaims.
Moreover, \percentage{77} of these top-\percentage{20} are the same \statistic{5} claims.
This reflects a property of the data: there are a few claims which are relevant to many motions, and are also implicitly mentioned in most speeches. 
Detection algorithms should be aware of this property, and account for it when evaluating performance.
At the same time, a claim's acceptance prior can be useful for inference. For example, it could be combined with other data in a more sophisticated algorithm, or could direct the parameter choice of such an algorithm.

\input{figures/precision_recall_on_test.tex}

%% file: figures/precision_recall_compare_to_idebate.tex
\begin{figure}[t]
\centering
\includegraphics[trim={2.3cm 8.9cm 2.1cm 8.7cm},clip,width=75mm]{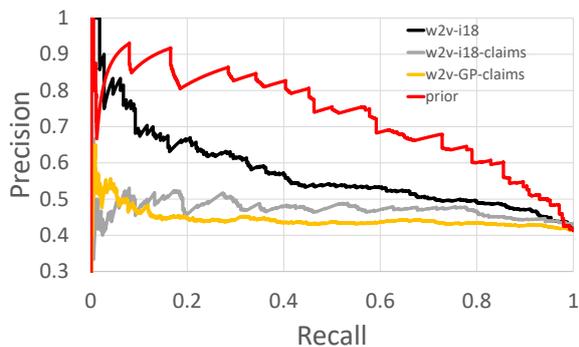}
\caption{
Precision-Recall curves for the matching of \CRKBclaims and \idebate claims to all $200$ speeches.
\label{fig:precision-recall-comparison-to-idebate}
}
\end{figure}

%% file: figures/precision_recall_on_test.tex
\begin{figure}[t]
\centering
\includegraphics[trim={2cm 4cm 2cm 3.5cm},clip,width=75mm]{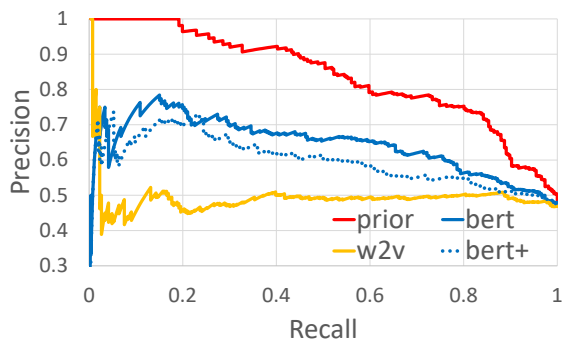}
\caption{Precision-Recall curves for matching \CRKBclaims to speeches, for \bertTestName (\percentage{20} of motions).
\label{fig:precision_recall_ont_test}
}
\end{figure}

%% file: Conclusions.tex
\section{Conclusions and Future Work}
We presented the problem of producing a rebuttal response to a long argumentative text.
This task is especially challenging when the discussed topic
is not known in advance, and, accordingly, potential responses are not readily available.

Toward the goal of addressing this problem we constructed a multi-layered dataset: (i) A Knowledge base of \CRKBclaims and corresponding rebuttal arguments, which are shown to be applicable for a wide variety of topics; (ii) A mapping of these claims to motions of \idebateDatasetName
in which they might be applicable; (iii) An annotation of the stance of applicable claims; (iv) An annotation for which claims are actually mentioned in relevant speeches, and whether they are mentioned explicitly or implicitly; (v) For explicitly mentioned claims, a (partial) annotation of which sentences imply them and which do not.

In addition, we presented 
baselines for the related Listening Comprehension task, suggesting that this is a complicated problem. Using state-of-the-art sentence embedding yielded an $F1$-score of $0.64$, while trivially taking the claim with the highest prior to be mentioned scored
$0.78$\footnote{Scores computed at the precision-recall intersection.}. This suggests that careful evaluation is required.

While baselines are provided only for 
detecting \CRKBclaims in spoken content, 
future work should aim to solve the problem as a whole - either by developing algorithms that determine relevance and stance of \CRKBclaims to given motions, or by forgoing these stages, and successfully deciding whether a claim was mentioned in a speech, without first focusing on relevant claims.

Our results suggest that GP rebuttal arguments usually work well as a response to speeches in which the matching claim was mentioned. However, this is by no means perfect; and in some \percentage{13} of the cases they do not. It is interesting to further identify and understand these cases. By doing so, an automatic system could prefer responses that it identifies as more appropriate. Moreover, understanding such cases can lead to improving the rebuttal arguments themselves.